\documentclass{article}
\usepackage{ijcai19}
\usepackage[english]{babel}

\usepackage{times}
\usepackage{soul}
\usepackage{url}
\usepackage[utf8]{inputenc}
\usepackage[small]{caption}
\usepackage{graphicx}
\usepackage{amsmath}
\usepackage{booktabs}
\title{Survey of Personalization Techniques for Federated Learning}

\author{
Viraj Kulkarni$^{1,2}$\and
Milind Kulkarni$^1$\and
Aniruddha Pant$^2$
\affiliations
$^1$Vishwakarma University\\
$^2$DeepTek Inc\\
\emails
}

\begin{document}

\maketitle

\begin{abstract}
Federated learning enables machine learning models to learn from private decentralized data without compromising privacy. The standard formulation of federated learning produces one shared model for all clients. Statistical heterogeneity due to non-IID distribution of data across devices often leads to scenarios where, for some clients, the local models trained solely on their private data perform better than the global shared model thus taking away their incentive to participate in the process. Several techniques have been proposed to personalize global models to work better for individual clients. This paper highlights the need for personalization and surveys recent research on this topic.
\end{abstract}

\section{Introduction}
Many datasets are inherently decentralized in nature and are distributed across multiple devices owned by different users. Traditional machine learning settings involve aggregating data samples from these users into a central repository and training a machine learning model on it. This movement of data from local devices to a central repository poses two key challenges. Firstly, it compromises the privacy and security of the data. Policies such as the General Data Protection Regulation (GDPR) \cite{voigt2017eu} and Health Insurance Portability and Accountability Act (HIPAA) \cite{annas2003hipaa} stipulate provisions that make such movement difficult. Secondly, it imposes communication overheads which, depending on the setting, may be prohibitively expensive.

Federated learning \cite{mcmahan2016communication} is a framework that enables multiple users known as \textit{clients} to collaboratively train a shared global model on their collective data without moving the data from their local devices. A central server orchestrates the federated learning process which consists of multiple rounds. At the beginning of each round, the server sends the current global model to the participating clients. Each client trains the model on its local data and communicates only the model updates back to the server. The server collects these updates from all clients and makes a single update to the global model thereby concluding the round. By removing the need to aggregate all data on a single device, federated learning overcomes the privacy and communication challenges mentioned above and allows machine learning models to learn on decentralized data.

Federated learning has found numerous practical applications where data is decentralized and privacy is important. For example, it has exhibited good performance and robustness for the problem of next-word-prediction on mobile devices \cite{hard2018federated}. Bonawitz et. al. \cite{bonawitz2019towards} propose a scalable system implementing large-scale federated learning for mobile devices. Kairouz et. al. \cite{kairouz2019advances} discuss broad challenges and open problems in the field.

The primary incentive for clients to participate in federated learning is obtaining better models. Clients who have insufficient private data to develop accurate local models stand to benefit the most from collaboratively learned models. However, the benefit of participating in federated learning for clients who have sufficient private data to train accurate local models is disputable. Yu et al. \cite{yu2020salvaging} show that, for many tasks, some participants may gain no benefit by participating since the global shared model is less accurate than the local models they can train on their own. Hanzely et al. \cite{hanzely2020federated} question the utility of a global model that is too far removed from the typical usage of a user. The distribution of data across clients is highly non-IID for many applications. This statistical heterogeneity makes it difficult to train a single model that will work well for all clients. The purpose of this paper is to survey recent research regarding building personalized models for clients in a federated learning setting that are expected to work better than the global shared model or the local individual models.

\section{Need for Personalization}
Wu et al. \cite{wu2020personalized} list three challenges faced by federated learning systems related to personalization: (1) device heterogeneity in terms of storage, computation, and communication capabilities; (2) data heterogeneity arising due to non-IID distribution of data; (3) model heterogeneity arising from situations where different clients need models specifically customized to their environment. As an example of model heterogeneity, consider the sentence: “I live in .....”. The next-word-prediction task applied on this sentence needs to predict a different answer customized for each user. If heterogeneity does not exist in the data, it may exist in the labels; different clients may assign different labels to the same data.

In the original federated learning design of McMahan et al. \cite{mcmahan2016communication}, the model updates and the final model can leak participant data violating privacy \cite{shokri2017membership} \cite{melis2019exploiting}. To preserve privacy, McMahan et al. \cite{mcmahan2017learning} propose differential privacy techniques that limit the information the global model can reveal about individual participants. However, Yu et al. \cite{yu2020salvaging} argue that such privacy protection mechanisms introduce a fundamental conflict between protecting privacy and achieving higher performance for individual users. Bagdasaryan et al. \cite{bagdasaryan2019differential} state that the cost of differential privacy mechanisms is the reduction in accuracy, and this cost is borne unequally by clients with the underrepresented or tail participants being affected the worst.

Personalization of the global model becomes necessary to handle the challenges posed by statistical heterogeneity and non-IID distribution of data. Most techniques for personalization \cite{sim2019investigation} generally involve two discrete steps. In the first step, a global model is built in a collaborative fashion. In the second step, the global model is personalized for each client using the client’s private data. Jiang et al. \cite{jiang2019improving} argue that optimizing solely for global accuracy yields models that are harder to personalize and propose that, in order to make federated learning personalization useful in practice, the three following objectives must all be addressed \textit{simultaneously} and not independently: (1) developing improved personalized models that benefit a large majority of clients; (2) developing an accurate global model that benefits clients who have limited private data for personalization; (3) attaining fast model convergence in a small number of training rounds. Out of the local data samples stored with each client, it may happen that only a subset of samples are relevant for a particular task, while the irrelevant samples adversely affect the model training. Tuor et al. \cite{tuor2020data} propose a method where a relevance model built on a small benchmark set is used to separate relevant and irrelevant samples at each client, and only the relevant samples are used in the federated learning process.

\section{Techniques}
This section surveys different methods for adapting global models for individual clients.

\subsection{Adding User Context}
Before presenting methods to personalize a global model for individual clients, we take a moment to point out that a shared global model can also generate highly personalized predictions if the client’s context and personal information is suitably featurized and incorporated in the dataset. However, most public datasets do not contain contextual features, and developing techniques to effectively incorporate context remains an important open problem that has great potential to improve the utility of federated learning models \cite{kairouz2019advances}. It also remains to be studied if such context featurization can be performed without adversely affecting privacy. As an intermediate approach between a single global model and purely local models, Masour et al. \cite{mansour2020three} suggest \textit{user clustering} where similar clients are grouped together and a separate model is trained for each group.

\subsection{Transfer Learning}
Transfer learning \cite{pratt1993discriminability} enables deep learning models to utilize the knowledge gained in solving one problem to solve another related problem. Schneider and Vlachos \cite{schneider2019mass} discuss using transfer learning to achieve model personalization in non-federated settings. Transfer learning has also been used in federated settings, e.g. Wang et al. \cite{wang2019federated}, where some or all parameters of a trained global model are re-learned on local data. A learning-theoretic framework with generalization guarantees is provided in \cite{mansour2020three}. By using the parameters of the trained global model to initialize training on local data, transfer learning is able to take advantage of the knowledge extracted by the global model instead of learning it from scratch. To avoid the problem of catastrophic forgetting \cite{mccloskey1989catastrophic} \cite{french1999catastrophic}, care must be taken to not retrain the model for too long on local data. A variant technique freezes the base layers of the global model and retrains only the top layers on local data. Transfer learning is also known as fine-tuning, and it integrates well into the typical federated learning lifecycle.

\subsection{Multi-task Learning}
In multi-task learning \cite{caruana1997multitask}, multiple related tasks are solved simultaneously allowing the model to exploit commonalities and differences across the tasks by learning them jointly. Smith et al. \cite{smith2017federated} show that multi-task learning is a natural choice to build personalized federated models and develop the MOCHA algorithm for multi-task learning in the federated setting that tackles challenges related to communication, stragglers, and fault tolerance. One drawback of using multi-task learning in federated settings is that since it produces one model per task, it is essential that all clients participate in every round.

\subsection{Meta-Learning}
Meta-learning involves training on multiple learning tasks to generate highly-adaptable models that can further learn to solve new tasks with only a small number of training examples. Finn et al. \cite{finn2017model} propose a model-agnostic meta-learning (MAML) algorithm that is compatible with any model that is trained using gradient descent. MAML builds an internal representation generally suitable for multiple tasks, so that fine tuning the top layers for a new task can produce good results. MAML proceeds in two connected stages: meta-training and meta-testing. Meta-training builds the global model on multiple tasks, and meta-testing adapts the global model individually for separate tasks.

Jiang et al. \cite{jiang2019improving} point out that if we consider the federated learning process as meta-training and the personalization process as meta-testing, then Federated Averaging \cite{mcmahan2016communication} is very similar to Reptile \cite{nichol2018first}, a popular MAML algorithm. The authors also make the observation that careful fine-tuning can produce a global model with high accuracy that can be easily personalized, but naively optimizing for global accuracy can hurt the model’s ability for subsequent personalization. While other personalization approaches for federated learning treat development of the global model and its personalization as two distinct activities, Jiang et al. \cite{jiang2019improving} propose a modification to the Federated Averaging algorithm that allows both to be addressed simultaneously resulting in better personalized models.

A new formulation of the standard federated learning problem proposed by Fallah et al. \cite{fallah2020personalized} incorporates MAML and seeks to find a global model which performs well \textit{after} each user updates it with respect to its own loss function. In addition, they propose Per-FedAvg, a personalized variant of Federated Averaging, to solve the above-mentioned formulation. Khodak et al. \cite{khodak2019adaptive} propose ARUBA, a meta-learning algorithm inspired by online convex optimization, and demonstrate an improvement in performance by applying it to Federated Averaging. Chen et al. \cite{chen2018federated} present a federated meta-learning framework for building personalized recommendation models where both the algorithm and the model are parameterized and need to be optimized.

\subsection{Knowledge Distillation}
Caruana et al. \cite{caruana1997multitask} have demonstrated that it is possible to compress the knowledge of an ensemble of models into a single model which is easier to deploy. Knowledge distillation \cite{hinton2015distilling} further develops this idea and involves extracting the knowledge of a large \textit{teacher} network into a smaller \textit{student} network by having the student mimic the teacher. Overfitting poses a significant challenge during personalization, especially for clients whose local dataset is small. Yu et al. \cite{yu2020salvaging} propose that by treating the global federated model as the teacher and the personalized model as the student, the effects of overfitting during personalization can be mitigated. Li et al. \cite{li2019fedmd} propose FedMD, a federated learning framework based on knowledge distillation and transfer learning that allows clients to independently design their own networks using their local private datasets and a global public dataset.

\subsection{Base + Personalization Layers}
In typical federated learning scenarios, data distribution varies greatly across participating devices. To temper the adverse effects of this statistical heterogeneity, Arivazhagan et al. \cite{arivazhagan2019federated} propose FedPer, a neural network architecture where the base layers are trained centrally by Federated Averaging, and the top layers (also called personalization layers) are trained locally with a variant of gradient descent. As opposed to transfer learning where all the layers are first trained on global data and then all or some layers are retrained on local data, FedPer separately trains the base layers on global data and the personalization layers on local data.

\subsection{Mixture of Global and Local Models}
The standard formulation of federated learning \cite{mcmahan2016communication} is designed to find a single global model trained on private data across all clients. Hanzely et al. \cite{hanzely2020federated} propose a different formulation of the problem that seeks an explicit trade-off between the global model and the local models. Instead of learning a single global model, each device learns a mixture of the global model and its own local model. To solve the formulation, the authors develop a new variant of gradient descent called Loopless Local Gradient Descent (LLGD). Instead of performing full averaging, LLGD only takes steps towards averaging thus suggesting that full averaging methods such as Federated Averaging might be too aggressive.

\section{Discussion}
Federated learning encompasses a wide variety of settings, devices, and datasets. When local datasets are small and the data distribution is IID, global models typically outperform local models, and a majority of clients benefit from participating in the federated learning process. However, when clients have sufficiently large private datasets and the data distribution is non-IID, local models exhibit better performance than the shared global model, and clients have no incentive to participate in the federated learning process. An open theoretical question is to determine the conditions under which shared global models can perform better than individual local models.

This paper surveys personalization techniques used to adapt a global federated model to individual clients. With a few exceptions, most prior work is focussed on measuring the performance of the global model on aggregated data instead of measuring its performance as seen by individual clients. Global performance, however, has no relevance if the global model is expected to be subsequently personalized before being put to use.

Personalized models usually show better performance for individual clients than global or local models. In some cases, however, personalized models fail to reach the same performance as local models, especially when differential privacy and robust aggregation is implemented \cite{yu2020salvaging}.

% \printbibliography 
\bibliographystyle{ieeetr}
\bibliography{bibliography}
\end{document}